\documentclass[sigconf]{acmart}

\AtBeginDocument{%
  \providecommand\BibTeX{{%
    \normalfont B\kern-0.5em{\scshape i\kern-0.25em b}\kern-0.8em\TeX}}}

\setcopyright{acmcopyright}

\copyrightyear{2022} 
\acmYear{2022} 
\setcopyright{acmcopyright}\acmConference[CIKM '22]{Proceedings of the 31st ACM International Conference on Information and Knowledge Management}{October 17--21, 2022}{Atlanta, GA, USA}
\acmBooktitle{Proceedings of the 31st ACM International Conference on Information and Knowledge Management (CIKM '22), October 17--21, 2022, Atlanta, GA, USA}
\acmPrice{15.00}
\acmDOI{10.1145/3511808.3557715}
\acmISBN{978-1-4503-9236-5/22/10}

\usepackage{algorithm}
\usepackage{algorithmicx}
\usepackage{algpseudocode}
\usepackage{makecell}
\usepackage{multirow}      
\usepackage{multicol} 
\usepackage{hyperref}
\floatname{algorithm}{Algorithm}

\usepackage{amsmath}
\usepackage{xcolor}
\usepackage{threeparttable}
\usepackage{subfigure}      
\usepackage{balance}

\begin{document}

\title{Trusted Media Challenge Dataset and User Study}

\author{Weiling Chen}\authornote{W. Chen is now with Hyundai Motor Group Innovation Center in Singapore.}
\email{weiling@aisingapore.org}
\orcid{0000-0002-6969-2696 }
\affiliation{%
  \institution{AI Singapore}
  \country{Singapore}
  \postcode{117602}
}

\author{Sheng Lun Benjamin Chua}
\email{benjaminchua95@hotmail.com}
\affiliation{%
  \institution{AI Singapore}
  \country{Singapore}
  \postcode{117602}
}

\author{Stefan Winkler}\authornote{S. Winkler is now with ASUS Intelligent Cloud Services (AICS) and National University of Singapore.}
\email{winkler@nus.edu.sg}
\affiliation{%
  \institution{AI Singapore}
  \country{Singapore}
  \postcode{117602}
}

\author{See-Kiong Ng}
\email{seekiong@nus.edu.sg}
\affiliation{%
  \institution{AI Singapore}
  \country{Singapore}
  \postcode{117602}
}

\renewcommand{\shortauthors}{Chen et al.}

\begin{abstract}
The emergence of fake media that can be easily created by technology has the potential to generate potent misinformation causing harm to both society and individuals. To tackle the issue, we have organized the Trusted Media Challenge (TMC) to explore how Artificial Intelligence (AI) technologies could be leveraged to combat fake media. 
To enable further research,  we are releasing the dataset from the TMC, consists of 4,380 fake and 2,563 real videos, with various video and audio manipulation methods employed to produce different types of fake media. We have also carried out a user study to demonstrate the quality of the TMC dataset and to compare the performance of humans and AI models. The results show that the TMC dataset can fool human participants in many cases.
The TMC dataset is available for research purposes upon request via \href{mailto:tmc-dataset@aisingapore.org}{tmc-dataset@aisingapore.org}. 
\end{abstract}

\ccsdesc[500]{Computing methodologies~Computer vision tasks}
\ccsdesc[500]{Applied computing~Computer forensics}

\keywords{datasets; deepfake detection; computer vision}


\maketitle

\section{Introduction}


Deepfakes are synthetic media in which a person in an existing image or video is replaced with someone else's likeness \cite{wiki:deepfakes}. They are generated by face swapping or face reenactment using deep learning techniques.
These Deepfakes with high authenticity pose new threats and risks in the form of scams, fraud, disinformation, social manipulation, or celebrity porn. In particular, fake audio-visual media such as news and interviews can be used to spread misinformation that may cause social fissures.

This inspired us to organize the Trusted Media Challenge (TMC) \cite{aisgtmc}, a five-month long competition for AI enthusiasts and researchers from around the world, to design and test out AI models and solutions that can effectively detect fake media created by various state-of-the-art Deepfake technologies. In particular, we focused on the detection of audiovisual fake media, where either or both video and audio modalities may be modified.  For further research by the community, we are releasing the TMC dataset which includes 4,380 fake and 2,563 real videos that were part of the challenge.

 High-quality fake video clips may include elaborate manipulations in both video (i.e.\ image frames) and audio tracks.  We believe an \emph{audiovisual} Deepfake dataset such as ours is useful to the research community. Another important feature of the TMC dataset is its focus on Asian content and ethnicities, whereas other datasets usually have a majority of Caucasians. Given that many Deepfake detectors are sensitive to skin tones and facial features, the TMC dataset serves as an important supplement to existing datasets and can help researchers develop more robust (and potentially less biased) Deepfake detectors.\footnote{~While the TMC dataset has been intentionally designed for evaluating the performance of detection models on Asian demographics,  using the TMC dataset on its own to develop detection models may lead to potential bias towards Asians, just as using existing datasets focused on Caucasians or other ethnicities can lead to the corresponding biases \cite{fabbrizzi2021survey}.}

The remainder of the paper is organized as follows. Section 2 reviews the related work on existing Deepfake datasets and user studies. Section 3 presents details of the construction of the TMC dataset. Section 4 introduces the experimental setting and results of the user study. Section 5 introduces the Trusted Media Challenge and the winning models. 
The last section concludes the  paper.

\section{Related Work}

\subsection{Deepfake Datasets}
Building large datasets of high quality for fake media detection requires much effort on data collection and media manipulation. There are some existing Deepfake datasets focusing on video manipulation, including FaceForensics++ (FF++) \cite{rossler2019faceforensics++}, DFD \cite{googledfd}, Celeb-DF \cite{li2020celeb}, DF-1.0 \cite{jiang2020deeperforensics}, DFDC \cite{dolhansky2020Deepfake}, WildDeepfake \cite{zi2020wilddeepfake} and KoDF \cite{kwon2021kodf}. Except for DFDC, all the other datasets only include video clips without audio. Even for the DFDC dataset, although the authors claim they have done some manipulations on audios, details are not disclosed and such manipulations are not actually leveraged in the challenge. In addition, only DFD and DFDC datasets have explicit consent from the actors in the videos. Moreover, the ethnicity distribution is not clear in many datasets, but it seems most of them have a majority of Caucasians and lack Asian subjects (e.g.\ 5.1\% in celeb-DF and 12\% in DFDC preview dataset \cite{dolhansky2019Deepfake}). One exception is KoDF \cite{kwon2021kodf}, a deepfake dataset focusing only on Koreans. 

There are far fewer existing datasets for fake audio detection. One of the most popular ones is the series of ASVspoof datasets \cite{wu2015asvspoof,kinnunen2017asvspoof,todisco2019asvspoof}. Among them, only ASVspoof 2015 specifically focused on the detection of synthetic and converted speech. 

A summary of the different datasets is shown in Table \ref{tab:table1}.

\subsection{User Studies}
Although there are numerous face forgery datasets, there are not many user studies on these datasets. FaceForensics++ conducted a user study with 204 participants (mostly computer science  students) on their dataset. The authors randomly set a time limit of 2, 4, or 6 seconds and then asked the attendees whether the displayed image selected from their test set was ‘real’ or ‘fake’. Each attendee was asked to rate 60 images, which resulted in a collection of 12,240 human decisions. 

Deep Video Portraits \cite{kim2018deep} and DeeperForensics \cite{jiang2020deeperforensics} designed a user study on real/fake videos.  100 professionals who specialized in computer vision research were engaged. Participants were then required to rate each clip on a scale from 1 to 5 according to its ``realness''. The authors assume that participants who give a score of 4 or 5 believe the video is real.

Again these user studies focused on human performance on the detection of fake videos rather than fake audio detection in the audiovisual setting. To fill this gap, we designed our user study that considers both video and audio modalities. 

\section{TMC dataset}
\subsection{Data sources}
We collected real-life footage to construct the TMC dataset from several different sources, including: 
\begin{itemize}
\item CNA\footnote{~CNA: https://www.channelnewsasia.com/} (Presenters and journalists talking in news programs)
\item ST\footnote{~ST: https://www.straitstimes.com/} (Interviewees answering questions in TV interviews)
\item Freelancers (people talking about different topics like hobbies, movies, food, etc.)
\end{itemize}

We obtained authorization to use and edit the footage produced by CNA and ST to generate and use fake clips for the purposes of the challenge. Similarly, we obtained consent from the freelancers to use and edit such clips to generate fake media. All the videos obtained from different sources are converted to mp4 format, and the resolution is changed to 360p or 1080p. More details about the collection and pre-processing of the video footage collected from the 3 sources are provided in Appendix \ref{sec:collection}.

\subsection{Generation Methods}
Based on the real videos collected from the above three sources, the fake media are created using different fake video and/or audio generation methods. 

\subsubsection{Video Manipulation}
The TMC dataset includes several fake video generation methods such as Deepfakes \cite{rossler2019faceforensics++}, FSGAN \cite{nirkin2019fsgan} and FaceSwap \cite{rossler2019faceforensics++}. We chose these methods as they were the most popular video manipulation methods at the time the dataset was created. 

\subsubsection{Audio manipulation}
We used StarGAN-VC \cite{kameoka2018stargan} and One-Shot VAE \cite{chou2019one} as audio manipulation methods to generate fake audio for the TMC dataset. 
A brief introduction of each video and audio generation method is provided in Appendix \ref{sec:generation}. 

\begin{table*}[htb]
\begin{threeparttable}[b]
\caption{Deepfake datasets}
\label{tab:table1}

\begin{tabular}{rrrrrrrr}
\hline
                              & Year     & Total                           & Consented                     & Video                       & Audio                       & Video                        & Audio                       \\
Dataset                       & released & clips                           & subjects\tnote{1}                      & methods\tnote{2}                     & methods                     & perturbs\tnote{3}                     & perturbs                    \\ \hline
FF++                          & 2019     & 5,000                           & NA                            & 4                           & 0                           & 0                            & 0                           \\ 
DFD                           & 2019     & 3,000                           & 28                            & 5                           & 0                           & 0                            & 0                           \\ 
Celeb-DF                      & 2019     & 6,229                           & NA                            & 1                           & 0                           & 0                            & 0                           \\ 
DF-1.0                        & 2020     & 60,000                          & 100                           & 1                           & 0                           & 7                            & 0                           \\ 
DFDC                          & 2020     & 128,154                         & 960                           & 8                           & 1                           & 16                           & 0                           \\ 
WildDeepfake                          & 2020     & 7,314                         & NA                           & NA                           & 0                           & 0                           & 0                           \\ 
KoDF                          & 2021     & 62,166                         & 403                           & 6                           & 0                           & 0                           & 0                           \\ 
ASVspoof2015                  & 2015     & 16,375                          & NA                            & 0                           & 10                          & 0                            & 0                           \\ 
\textbf{TMC} & \textbf{2022}     & \textbf{6,943} & \textbf{181} & \textbf{4} & \textbf{2} & \textbf{12} & \textbf{3} \\ \hline
\end{tabular}
   \begin{tablenotes}
     \item[1] Number of subjects who have given their consent to use and edit their videos.
     \item[2] Number of fake video generation methods used.
     \item[3] Number of video perturbation methods applied.
   \end{tablenotes}
\end{threeparttable}
\end{table*}

\subsubsection{Lip synchronization error}
To create videos with a mismatch between video and audio content, we take the video footage from one real video and the audio segment from another real video and combine them into a new clip. However, we impose the restriction that both the audio and video come from persons with the same gender. 

\subsection{Perturbations}
To simulate transmission errors, compression as well as different visual and audio effects, we add perturbations to both video and audio tracks.  Detailed proportion of each perturbation method and sample frames are provided in Appendix \ref{app:perturbations}.

\subsubsection{Video Perturbations} 

We selected 12 video perturbation methods to apply on the TMC dataset, which can be divided into 3 categories:
\begin{itemize}
    \item Weather effects (fog, snow, etc.) 
    \item Lighting effects (brightness change, contrast change, etc.) 
    \item Others (compression, scaled, shaky camera, etc.) 
\end{itemize}
Lighting and shaky camera effects account for a slightly higher percentage compared to other perturbation methods. 

We utilize \textit{imgaug}\footnote{~imgaug: https://imgaug.readthedocs.io/en/latest/} to add perturbations to videos in the TMC dataset. For each original real or fake video which needs to be perturbed, we randomly create 0 to 4 copies and chose one perturbation method to apply on each copy. The parameters for each perturbation are randomly set within a controlled range. 

To avoid misuse of the TMC dataset, we have added a watermark to all videos. This is not regarded as a perturbation but may still have an impact on some fake detectors. 
 
\subsubsection{Audio Perturbations}
We selected 3 audio perturbation methods to apply on the TMC dataset, which are volume change, additive Gaussian noise, and frequency masking, in almost equal proportion. 
We utilize \textit{nlpaug}\footnote{~nlpaug: https://github.com/makcedward/nlpaug} to add perturbations to audios in the TMC dataset. Similar to video perturbations, parameters are initialized randomly within a controlled range.

\subsection{Dataset Content}
\label{S:DatasetContent}
There are four types of fake media in the TMC dataset, as  listed in Table \ref{tab:type}. The last column shows the percentage of each fake type in the TMC dataset. Type-4 represents lip sync error described above -- although both video and audio are real, the speech content does not match the mouth movement in the videos. Therefore, it is regarded as fake as well.

\begin{table}[htb]
\caption{Different types of fake media in the TMC dataset.}
\label{tab:type}
\centering
\begin{tabular}{cccr}
\hline
Type & Video & Audio & Percentage \\ \hline
1             & Fake           & Fake           & 20.24\%                    \\ 
2             & Fake           & Real           & 22.97\%                    \\ 
3             & Real           & Fake           & 9.07\%                    \\ 
4             & Real           & Real           & 10.80\%                    \\ \hline
Real          & Real           & Real           & 36.92\%                    \\ \hline
\end{tabular}
\end{table}


The training dataset for the challenge did not include the detailed labels of different fake types but such labels are available together with the TMC dataset upon request. 

The TMC dataset includes a training set and two hidden test sets. The training set is released together with the challenge. 
It consists of 4,380 fake and 2,563 real videos sourced from 181 subjects in total. As mentioned, the TMC dataset differs from many other public datasets as it focuses on the Asian ethnic group, with 72.65\% of subjects Asians 
and 45.82\% of female subjects. 
The videos have a variety of durations, with a minimum length of 10 seconds. Additionally, both high (1080p) and low (360p) resolution videos are included in the training set. 


The test set is kept hidden to evaluate the performance of the detection models submitted as part of the challenge. The challenge was conducted in two phases, with a separate test set for each phase. The ratio of real and fake videos of both test sets are similar to that of training set. However, the test set in Phase II (2,950 videos) is larger than that of Phase I (1,000 videos) and also richer in distribution compared to the training set and the test set used in Phase I. This was done to better simulate real-world scenarios for Deepfake detection.

\section{User Study}
\label{sec:user_studies}

Our user study design was inspired by \cite{rossler2019faceforensics++,jiang2020deeperforensics}.  We randomly selected 5,000 videos from the TMC dataset, FaceForensics++, Deeper-Forensics-1.0 as well as DFDC dataset, with a proportion of 70\%, 10\%, 10\%, 10\% respectively. Each video in the user study dataset was viewed by at least 3 participants. 

We recruited 176 participants, most of whom were university students. Each participant viewed and rated a set of 100 videos with various distributions of real/fake videos.  For each video, participants were given 6, 8 or 10 seconds to view/listen before being redirected to the rating page. For videos sampled from the TMC dataset, which have both video and audio tracks, the participants were asked to give feedback via a 5-level Likert scale (strongly agree, agree, unsure, disagree, strongly disagree) to the following  statements:
\begin{itemize}
    \item This video looks real to me.
    \item This audio sounds real to me.
    \item Overall, this clip seems real to me.
\end{itemize}
For a video-only clip sampled from the other datasets, the participants were only asked to give feedback on video. 

\subsection{Comparisons with Other Datasets}
The results of the user study are shown in Table \ref{tab:table4}.  We assume that participants who indicated "strongly agree" or "agree" believed the video/audio is real, as also shown in the table. Since other datasets only include video manipulation, we only consider Type-1 and Type-2 fake media in TMC for comparisons in this table. 

\begin{table}[htb]
\centering
\begin{threeparttable}
\caption{Distribution of user ratings for fake videos in different datasets.}
\label{tab:table4}
\begin{tabular}{rrrrrr|r}
\hline
 & strongly & & & &  strongly \\
 & agree & agree & unsure & disagree &  disagree & ``real''\\ \hline
DF-1.0                 & 0.007                    & 0.067                    & 0.073                    & 0.234                    & 0.619                    & 0.074                         \\ 
DFDC                & 0.298                    & 0.348                    & 0.074                    & 0.136                    & 0.144                    & 0.646                         \\ 
FF++                  & 0.005                    & 0.026                    & 0.042                    & 0.166                    & 0.760                    & 0.031                         \\ 
\textbf{TMC} & \textbf{0.054}                    & \textbf{0.146}                    & \textbf{0.085}                    & \textbf{0.239}                    & \textbf{0.476}                    & \textbf{0.200}                         \\ \hline
\end{tabular}
\end{threeparttable}
\end{table}

It can be observed from Table \ref{tab:table4} that fake videos in TMC look more realistic than those in FF++ and DF-1.0, but not as realistic as those in DFDC. One possible reason is that the TMC dataset has different fake video types. For example, other datasets do not contain Type-4 (i.e.\ inconsistency in lip movement and speech content) fakes, which is very easy for people to spot (see Section \ref{sec4.3}). 

To understand whether the proportion of fake videos would affect human performance in detecting fake videos, the set of 100 videos viewed by each participant had a random distribution of real/fake videos sampled from 1:99, 10:90, 20:80, 30:70, …, 90:10, 99:1. To compare human performance with that of AI models, we use Area Under the Curve (AUC) to denote human performance. Participants' ratings - Strongly agree, agree, unsure, disagree, strongly disagree were converted to probabilities of 0, 0.25, 0.5, 0.75, 1 respectively to calculate AUC. 

The AUC of human detection for different ratios of real/fake videos ranges from 0.840 to 0.895 without obvious trends. This indicates that the proportion of fake videos does not affect human judgement significantly. In comparison, the best AI model submitted for the Challenge achieved an AUC of 0.9853.


\subsection{Exploring Generation Methods, Fake Types and others}
\label{sec4.3}
We have further summarized participants' performance in detecting fake media in terms of different generation methods and fake types. Results are shown in Tables \ref{tab:table5} and \ref{tab:table6}.

\begin{table}[htb]
\caption{User performance in detecting different types of fake media}
\label{tab:table5}
\centering
\begin{tabular}{lc}
\hline
Fake Type & AUC score \\ \hline
Type-1             & 0.914              \\ 
Type-2             & 0.864              \\ 
Type-3             & 0.769              \\ 
Type-4             & 0.930               \\ 
Overall             & 0.870               \\ \hline
\end{tabular}
\end{table}

\begin{table}[htb]
\caption{User performance in detecting fake media generated by different methods}
\label{tab:table6}
\centering
\begin{tabular}{lc}
\hline
Method & AUC score \\ \hline
Deepfakes-256             & 0.890               \\ 
Deepfakes-512             & 0.810               \\ 
Faceswap                  & 0.943              \\ 
FSGAN                     & 0.915              \\ \hline
One-Shot VAE              & 0.930               \\ 
StarGAN-VC                & 0.849              \\ \hline
\end{tabular}
\end{table}

It can be observed from the tables that participants can easily spot mismatched audio and video, making Type-4 the easiest type to be detected, followed by Type-1 in which both videos and audios are manipulated -- presumably, as long as humans can detect some artifacts in either audio or video track, they would label the video as fake.  Type-3 is the most difficult type for participants to detect, which indicates that people are more sensitive to video manipulation than audio manipulation. 

In terms of video manipulation methods, Deepfakes-512 produces the most realistic videos. For audio manipulation, StarGAN-VC performs better than One-Shot VAE. In addition, the sources of source and target videos also have an impact on human performance. If both videos come from the same source, it is harder for participants to detect the fake videos (AUC 0.74), while it is easier if the videos are from different sources (AUC 0.79). 

Different perturbations have similar impact on human performance. However, more videos are rated as fake regardless of their labels. This indicates that videos are more likely to be rated as fake when they are distorted in some way, but it does not become easier for participants to detect fake videos with perturbations present. A likely explanation for this finding is that some participants interpret perturbations as features of fake media. 

\section{Trusted Media Challenge}

The Trusted Media  Challenge was hosted by AI Singapore from July to December 2021 with a total prize pool of up to SGD 700K (approximately USD 500K). In total, there were 589 registered participants forming 475 teams registered for the challenge. The challenge consisted of two phases with different test sets, as described in Section \ref{S:DatasetContent}.  

The 3 winning solutions are based on modular detection of the different types of manipulations presented in the competition including video manipulation, audio manipulation as well as inconsistency in lip movement and speech content. The final result was calculated by combining individual detection scores from different models using specific aggregation strategies.
The best team achieved an AUC of 0.9853. Further details of the top 3 models and their performance are  provided in the Appendix \ref{TMC}.

\section{Conclusions}
In this work, we presented a fake media dataset consisting of 6,943 fake and real videos generated with a variety of manipulation methods. Different perturbations were added to both real and fake videos to increase the difficulty of detection. The dataset was used in the Trusted Media Challenge held in 2021 by AI Singapore. 

We have designed and carried out a comprehensive user study to demonstrate the quality of TMC dataset and explored the impact on human performance with regards to different factors. Our study results indicated that the TMC dataset can fool humans in many cases.  However, based on the results from the Trusted Media Challenge,  we also found that AI models can beat humans for Deepfake detection on our TMC dataset. 


\begin{acks}
This research/project is supported by the National Research Foundation, Singapore under its AI Singapore Programme. 
\end{acks}

\newpage
\balance
\bibliographystyle{ACM-Reference-Format}  


\appendix
\section{Details of the TMC Dataset}
\subsection{Collection and Pre-processing}
\label{sec:collection}
For videos provided by CNA and ST, we extract footage which includes talking persons using a combination method of face detection and human labeling. The reason for this extraction phase is that videos from the two aforementioned sources  include not only talking persons but also many voice-over scenes. 

Freelancers were given detailed instructions on how to take selfie videos used for the TMC dataset. They have to ensure that their face is recognisable such that the head is between 1/8 and 1/2 of the video’s total height from the bottom of the chin to the top of the head and is unobstructed by objects such as hair or face masks, so that key facial features like eyes, eye brows, nose, mouth and chin are clearly visible. Each freelancer was asked to create 5 videos, out of which: 
\begin{itemize}
    \item At least 1 must be taken indoors (e.g.\ home, office, coffee shop), and at least 1 outdoors (e.g.\ street, playground, park);
    \item At least 1 must be taken with back lighting, at least 1 with frontal lighting;
    \item At least 1 must be taken in the daytime, at least 1  in the evening or at night;
    \item At least 3 with the face directly facing the camera.
\end{itemize}

Each video must have a duration of at least 45 seconds (recommended length 45-60 seconds), a minimum resolution of 1080p and a minimum frame rate of 24fps. Each audio must have a minimum of 32 bits per sample and a minimum sampling rate of 48kHz. The freelancer's voice must be clearly heard, and background noise be kept to a minimum.  

After the freelancers submitted their videos, the freelancing platform and our team checked the content of the videos to make sure the above requirements were met before including them for further processing. 

\begin{figure*}[htbp]
\centering
\subfigure[Different head sizes.]{
\begin{minipage}[t]{0.25\linewidth}
\centering
\includegraphics[width=1.7in]{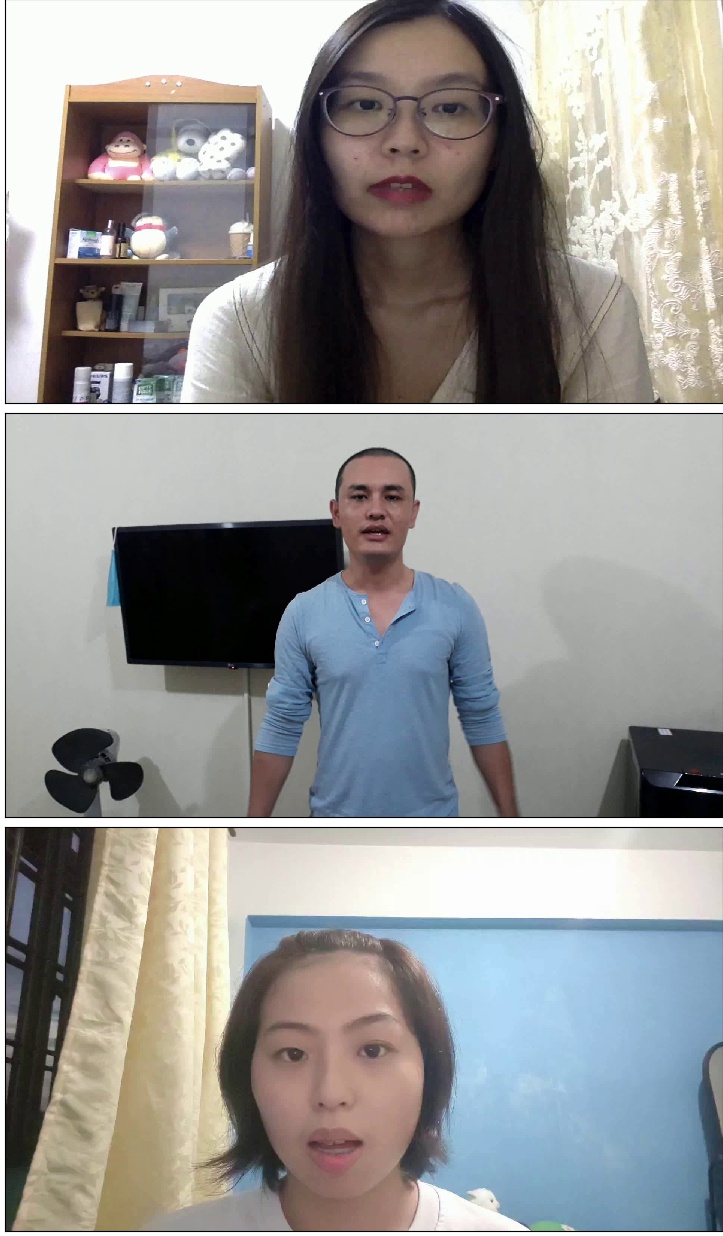}
\end{minipage}%
}%
\subfigure[Different face angles.]{
\begin{minipage}[t]{0.25\linewidth}
\centering
\includegraphics[width=1.7in]{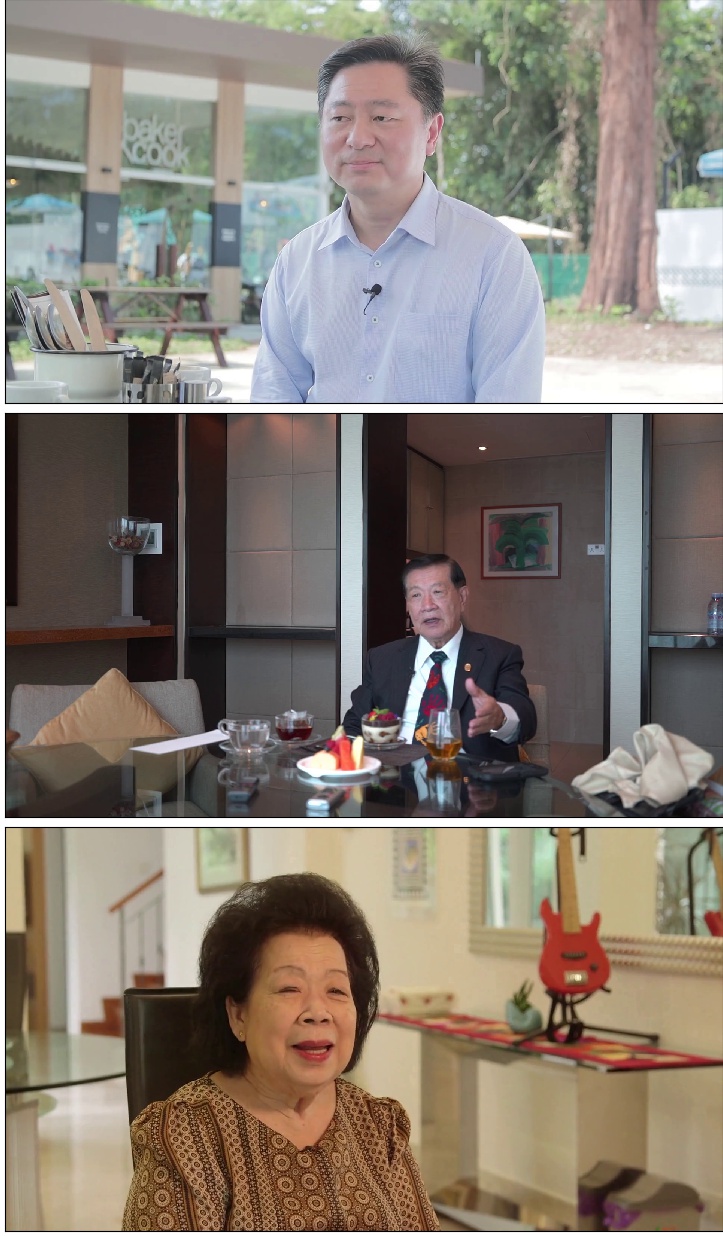}
\end{minipage}%
}%
\subfigure[Different lighting.]{
\begin{minipage}[t]{0.25\linewidth}
\centering
\includegraphics[width=1.7in]{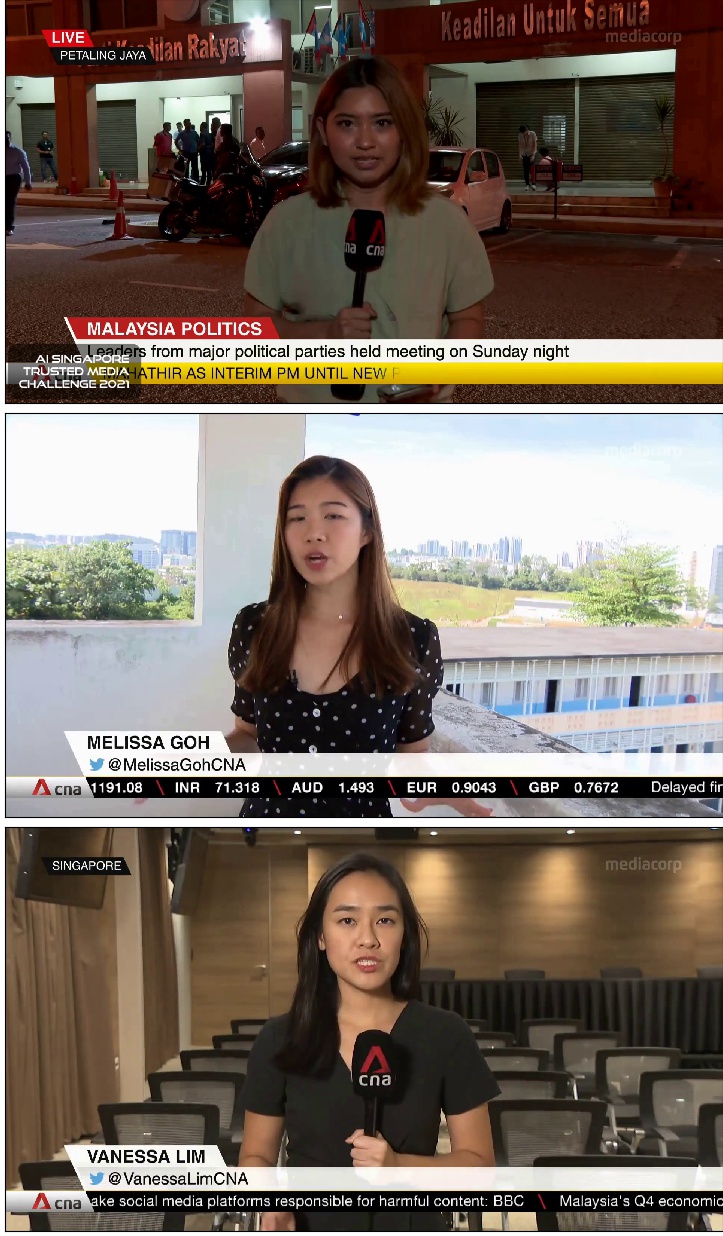}
\end{minipage}
}%
\subfigure[Multiple faces.]{
\begin{minipage}[t]{0.25\linewidth}
\centering
\includegraphics[width=1.7in]{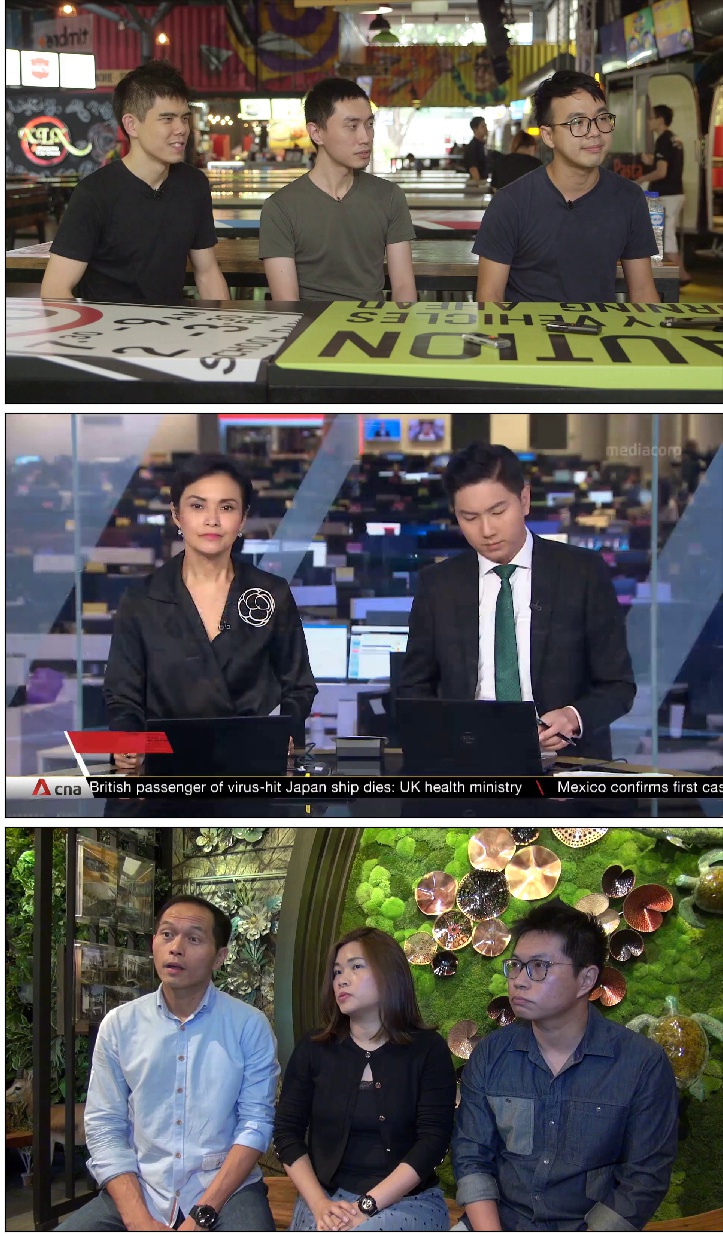}
\end{minipage}
}%
\centering
\caption{Diversity in the TMC dataset.}
\label{fig:sample}
\end{figure*}

Figure \ref{fig:sample} demonstrates the diversity of the dataset with frames from several sample videos in the TMC dataset. 

\subsection{Generation Methods}
\label{sec:generation}
This subsection gives a brief introduction of each video and audio manipulation methods we have applied to generate TMC dataset.

\subsubsection{Video Manipulation}
The TMC dataset includes several fake video generation methods. We chose these methods as they were the most popular video manipulation methods at the time the dataset was created. A brief introduction of each method is provided below. 

\textbf{Deepfakes-256/512}: 
Deepfakes is also the name of a popular video manipulation method. 
In the training phase, sets of source and target faces are fed into a shared encoder to learn a single model for two different people. Meanwhile, two decoders are trained to reconstruct the two sets of faces. A face detector is applied to crop and align the images. We set the output size of extracted faces to 256x256 and 512x512 to produce fake videos of different quality. In the inference phase, the two sets of faces run through the opposite encoder so as to produce a realistic swap. 
For TMC dataset, we use the github implementation of Deepfakes faceswap\footnote{~Deepfakes faceswap: https://github.com/Deepfakes/faceswap}.  

\textbf{FSGAN}: We created FSGAN fake videos using the pretrained models provided in \cite{nirkin2019fsgan}.  Unlike previous Deepfake generation methods, FSGAN is subject agnostic, meaning it can be applied directly to source-target face pairs without training on them. FSGAN includes a recurrent reenactment generator and segmentation generator which analyze the source and target faces respectively, from which the inpainting generator creates the reenacted face. A blending generator is applied for the final fake images/videos. 

\textbf{FaceSwap} is a classical computer graphics-based method. The face region and its landmarks are extracted for the source face. The 3D model is fit to the landmarks and then blended with the images of the target face using feathering mechanism and color correction.
For the TMC dataset, we created FaceSwap videos using Marek Kowalski's implementation on github\footnote{~FaceSwap: https://github.com/MarekKowalski/FaceSwap/}. 

\textbf{Other methods}: Two additional video manipulation methods not used in the released TMC training dataset were used to generate fake videos for the Phase II test set of the challenge. Without disclosing too many details, they are face swap and face re-enactment methods as well. These videos account for less than 10\% of the total videos in the Phase II test set.


In a face swap scenario, the goal is to generate a fake video of the source person. Therefore, the face of the source person is ``extracted'' from the source video and ``replaces'' the face of the target person in the target video. To make the best of existing fake video generation methods, it is important to select suitable source-target pairs to generate more deceptive fake media. In this matching phase, we consider several attributes, including gender, skin tone, face shape, hair covering (i.e.\ face and forehead), and spectacles. We assign higher weights to features like gender and skin tone to ensure better matching. 
For each person, we manually label these attributes and try to match those with highest similarity for face swapping to ensure better quality of the generated fake media.

\subsubsection{Audio manipulation}
We used the following audio manipulation methods to generate fake audio for the TMC dataset. 

\textbf{StarGAN-VC}  \cite{kameoka2018stargan} is a popular fake audio generation method. It uses a generative adversarial network (GAN) called StarGAN as a base to enable non-parallel many-to-many voice conversion (VC). StarGAN-VC includes a generator which takes acoustic features of Person A's speech and the Person B's speaker identity label as inputs to generate acoustic features of speech from Person A with Person B's identity label. Other important components are the real/fake discriminator and the domain classifier, which is used to predict the speaker identity.  For the TMC dataset, we train our own models based on all the speakers together.

\textbf{One-Shot VAE} \cite{chou2019one} is an  approach which only requires an example utterance from source and target speaker respectively to perform voice conversion. No training for the source and target speaker is required. A speaker encoder is used to extract the speaker representation from the source person\footnote{~The definition of the source and target speaker in their paper is opposite to ours.} while a content encoder is used to extract content representation from the target utterance. The outputs of two encoders are then integrated and fed into the decoder to generate the converted voice.  For the TMC dataset, we use the official implementation from the paper and their pretrained models. 

\subsection{Perturbations}
\label{app:perturbations}
\begin{figure*}[htb]
  \centering
  \includegraphics[width=1\textwidth]{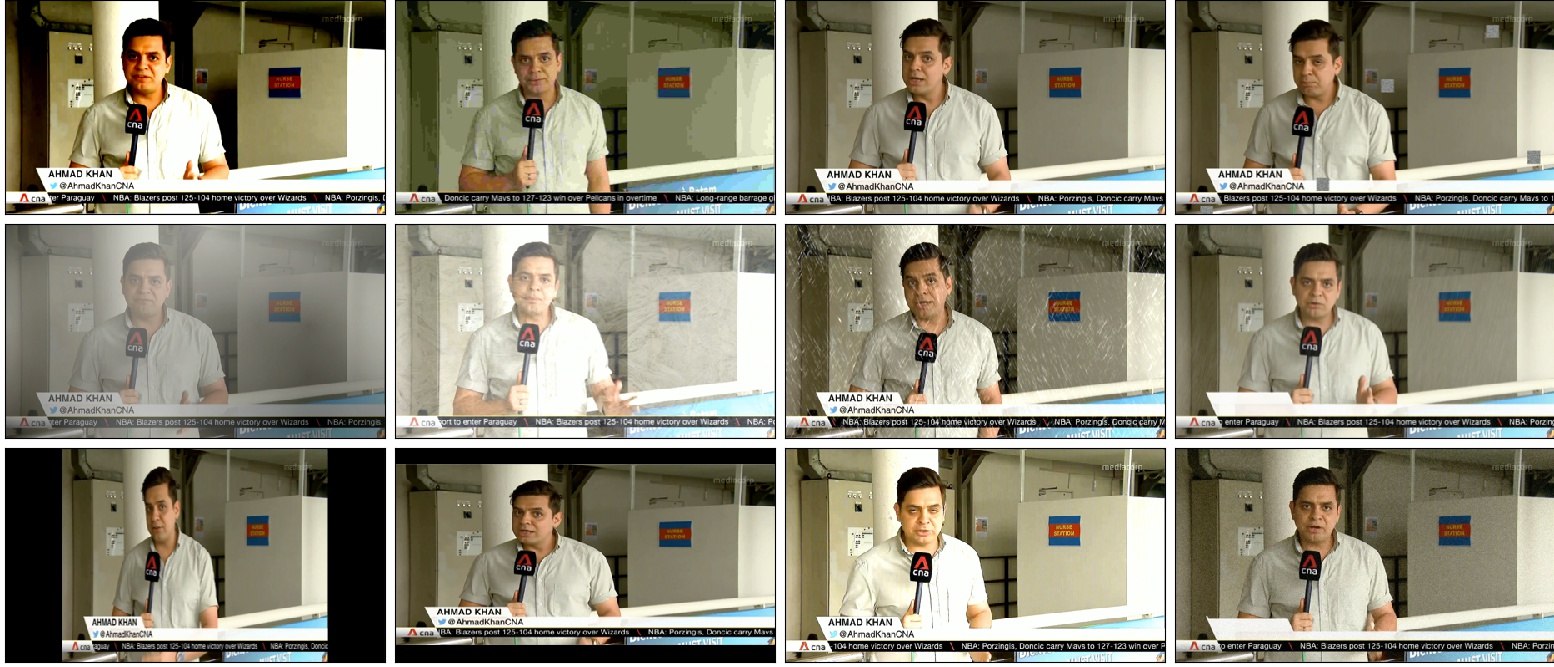}
  \caption{A series of frames from a sample video with different perturbations applied. Top row (from left to right): contrast and brightness change, compression, crop, cutout. Middle row: fog, frost, snow, rain. Bottom row: horizontal scaling, vertical scaling, sharpen, noise. }
  \label{fig_perturb}
\end{figure*}

We applied 12 video perturbation methods and 3 audio perturbation methods in the TMC dataset. 
Figure \ref{fig_perturb} shows frames from a sample video with different perturbations applied. Table \ref{tab:video_perturb} and \ref{tab:audio_perturb} shows the proportion of various video and audio perturbation methods respectively.

\begin{table}[htb]
\caption{Percentages of different video perturbations in the TMC dataset.}
\label{tab:video_perturb}
\centering
\begin{tabular}{lr}
\hline
Video          &            \\
perturbation   & Percentage \\ \hline

lighting        & 9.02\% \\
shaky           & 8.38\% \\
scaley         & 6.05\% \\
cutout         & 6.03\% \\
compression    & 5.89\% \\
snow           & 5.75\% \\
rain           & 5.66\% \\
fog            & 5.62\% \\
sharpen        & 5.57\% \\
noise          & 5.55\% \\
scalex         & 5.24\% \\ \hline
No perturbs       & 31.24\% \\ \hline
\end{tabular}
\end{table}

\begin{table}[htb]
\caption{Percentages of different audio perturbations in the TMC dataset.}
\label{tab:audio_perturb}
\centering
\begin{tabular}{lr}
\hline
Audio          &            \\
perturbation   & Percentage \\ \hline
volume change  & 11.18\%           \\
Gaussian noise & 11.58\%           \\
frequency mask & 11.35\%           \\ \hline
No perturbs    & 65.89\%           \\ \hline
\end{tabular}
\end{table}


\balance
\section{Trusted Media Challenge}
\label{TMC}
The Trusted Media  Challenge consisted of two phases with different test sets, as described in Section \ref{S:DatasetContent}.  Phase I lasted for four months while Phase II lasted for one month. Participants could submit their Deepfake detection models trained on the TMC Dataset with or without external datasets.  90 teams made successful submissions.  After Phase I, 15 teams qualified to enter Phase II to compete for the top 3 places.

\subsection{Top 3 Teams}
All 3 winning solutions are based on modular detection of the different types of manipulations presented in the competition including video manipulation, audio manipulation as well as inconsistency in lip movement and speech content.  The final scores and running time for the top 3 teams in Phase II are shown in Table \ref{tab:ranking}.

\begin{table}[htb]
\caption{Performance of top 3 solutions.}
\centering
\label{tab:ranking}
\begin{tabular}{cccc}
\hline
Ranking & Team  & AUC & Time(s) \\ \hline
1       & Will           & 0.9853       & 15102              \\ 
2       & IVRL           & 0.9833       & 25554              \\ 
3       & HideOnFakeBush & 0.9819       & 15989              \\ \hline
\end{tabular}
\end{table}


\subsubsection{Team Will (1st place)} The top team achieves the highest AUC with fastest inference speed in terms of running time. They divided  the problem into two sub-tasks:
\begin{itemize}
    \item To distinguish “either face or audio or both are manipulated” from real.
    \item To distinguish “lip motions inconsistent with the audio” from real.
\end{itemize}


The first classifier takes the face crop and audio clips as inputs. The audio is further converted into Mel-spectrogram. Both face crops and Mel-spectrogram images then go through a Convolutional Neural Network (CNN) encoder respectively. The two output feature vectors are forwarded to a fusion head as final stage, and the network would output a single neural unit as the prediction logit.

Apart from the  multi-modal audiovisual classifiers, the winning solution further utilized knowledge distillation \cite{hinton2015distilling} to boost the performance. This is the biggest difference between their solutions and other teams' solutions. They leveraged both single and multi-modal models and tried different architectures such as EfficientNet \cite{tan2019efficientnet}, Swin Transformers, ResNet \cite{he2016deep}, MobileNets \cite{howard2017mobilenets} to ensure the diversity of teachers. 

For lip movement inconsistency, Team Will observed that detection became easier if they categorized the input videos into broadcast TV or amateur selfie scenarios. Based on this observation, they first used a DNN scenario classifier, then used the joint audio-video manipulation detection model structure for detailed classification. 

\subsubsection{Team IVRL (2nd place)}

The solution from Team IVRL consecutively predicts the authenticity of the audio, face, and lip-sync with three models separately in a cascade arrangement to make it computationally efficient. 

For the face part, they used ensembled EfficientNet trained on DFDC dataset. Forthe audio part, they adapted a ResNet \cite{alzantot2019deep} designed for the Automatic Speaker Verification Spoofing and Countermeasures Competition (ASVspoof) as fake audio. For lip-sync, they used off-the-shelf method SyncNet \cite{chung2016out} which measured the coherence between a short audio clip (0.2 seconds) and a series of images of the talking person (five frames). 

\subsubsection{Team HideOnFakeBush (3rd place)}
The team chose an approach that was quite similar to Team IVRL. They also used 3 different models to detect audio manipulation (ResNet50), face manipulation (WSDAN \cite{hu2019see}) and lip-sync errors (SyncNet) respectively and aggregate at the score level using a random forest. 

\subsection{Analysis}
Similar to user studies described in Section \ref{sec:user_studies}, we further studied the performance of winners' models on different types of fake media and generation methods. The results are shown in the following tables.

\begin{table}[htb]
\caption{Top 3 teams' performance (AUC) in detecting different types of fake media.}
\centering
\label{tab:ai_type}
\begin{tabular}{llll}
\hline
        & Team 1 & Team 2 & Team 3 \\ \hline
Type-1  & 0.9989       & 0.9972       & 0.9959       \\ 
Type-2  & 0.9716       & 0.9616       & 0.9565       \\ 
Type-3  & 0.9981       & 0.9950       & 0.9964       \\ 
Type-4  & 0.9786       & 0.9891       & 0.9899       \\ \hline
Overall & 0.9853       & 0.9833       & 0.9819       \\ \hline
\end{tabular}
\end{table}

\begin{table}[htb]
\caption{Top 3 teams' performance (AUC) in detecting fake media generated by different methods.}
\centering
\label{tab:ai_method}
\begin{tabular}{llll}
\hline
Method & Team 1 & Team 2 & Team 3 \\ \hline
Deepfakes-256 & 0.9970       & 0.9955       & 0.9944       \\ 
Deepfakes-512 & 0.9749       & 0.9928       & 0.9929       \\ 
Faceswap      & 0.9846       & 0.9786       & 0.9805       \\ 
FSGAN         & 0.9956       & 0.9859       & 0.9891       \\ \hline
VAE           & 0.9992       & 0.9949       & 0.9966       \\ 
starGAN       & 0.9984       & 0.9963       & 0.9961       \\ \hline
\end{tabular}
\end{table}

Different from human performance, it can be observed from Table \ref{tab:ai_type} that Type-2 is the most difficult for AI models to detect, followed by Type-4. Type-1 is the easiest to be detected - same as for humans. 
Table \ref{tab:ai_method} shows that the top winning models have different performance on different video generation methods, while their performance is similar for audio generation methods. 
Overall, AI models' performance is much better than that of humans.  

\end{document}